\documentclass[10pt, conference]{IEEEtran}

\usepackage{amsmath,amssymb,amsfonts}
\usepackage{graphicx}
\usepackage{pifont}
\usepackage[utf8]{inputenc}

\usepackage{tikz}
\usepackage{tikz-cd}
\usepackage{pgfplots}
\pgfplotsset{compat=1.14}
\usepackage{cite}

\usepackage{lipsum}

\usepackage{amsthm}
\usepackage{xcolor}
\usepackage[figuresright]{rotating}

\usepackage{graphicx}
\graphicspath{{/}{fig/}}

\usepackage{array}
\usepackage{textcomp}
\usepackage{xcolor}
\usepackage{multirow}
\usepackage{booktabs}

\usepackage{mathtools}
\usepackage{breqn}
\usepackage{float}

\usepackage{pgfplots}
\pgfplotsset{compat=1.7}
\usepgfplotslibrary{groupplots}

\usepackage{caption}
\usepackage{subcaption}

\usepackage{multirow,tabularx}
\usepackage{hyperref}
\usepackage{flushend}
\usepackage{algorithmic}
\usepackage[vlined, ruled, shortend]{algorithm2e}

\newlength\figureheight
\newlength\figurewidth
\SetAlCapNameFnt{\footnotesize}
\SetAlCapFnt{\footnotesize}

\title{
    Follow-Me in Micro-Mobility with \\End-to-End Imitation Learning \\
}

\author{
    \IEEEauthorblockN{
        \vspace{1em}
        Sahar Salimpour\IEEEauthorrefmark{2},
        Iacopo Catalano\IEEEauthorrefmark{2}\IEEEauthorrefmark{3},
        Tomi Westerlund\IEEEauthorrefmark{2},
        Mohsen Falahi\IEEEauthorrefmark{3},
        Jorge Peña Queralta\IEEEauthorrefmark{3}\IEEEauthorrefmark{4}
    }
    \IEEEauthorblockA{
        \normalsize
        \IEEEauthorrefmark{2}\href{https://www.utu.fi/en/university/faculty-of-technology/computing}{Department of Computing, University of Turku, Finland}.\\
        \IEEEauthorrefmark{3}\href{https://daav.ch}{DAAV S.A., Switzerland}.\\
        \IEEEauthorrefmark{4}\href{https://www.zhaw.ch/en/engineering/institutes-centres/cai}{Centre for Artificial Intelligence, Zürich University of Applied Sciences, Switzerland}.\\
        Emails: \IEEEauthorrefmark{2}\{sahars, imcata, tovewe\}@utu.fi, \IEEEauthorrefmark{3}mohsen@daav.ch, \IEEEauthorrefmark{4}penq@zhaw.ch
        \\[+6pt]
    }
}

\begin{document}

\maketitle
\thispagestyle{empty}
\pagestyle{empty}



\begin{abstract}%
    \label{sec:abstract}%
    Autonomous micro-mobility platforms face challenges from the perspective of the typical deployment environment: large indoor spaces or urban areas that are potentially crowded and highly dynamic. While social navigation algorithms have progressed significantly, optimizing user comfort and overall user experience over other typical metrics in robotics (e.g., time or distance traveled) is understudied. Specifically, these metrics are critical in commercial applications. In this paper, we show how imitation learning delivers smoother and overall better controllers, versus previously used manually-tuned controllers. We demonstrate how DAAV’s autonomous wheelchair achieves state-of-the-art comfort in follow-me mode, in which it follows a human operator assisting persons with reduced mobility (PRM). This paper analyzes different neural network architectures for end-to-end control and demonstrates their usability in real-world production-level deployments.

\end{abstract}

\IEEEpeerreviewmaketitle


\section{Introduction}\label{sec:introduction}

Autonomous vehicles are gaining significance with advancements in robotics across diverse fields such as healthcare, transportation, and manufacturing. Extensive research has been conducted on autonomous navigation systems such as assistive human-following vehicles, designed to transport loads in public environments such as hospitals, restaurants, and airports. Various types of vehicles, ranging from wheeled mobile robots to UAVs~\cite{evangeliou2022visual}, have been used in such applications, where the leader's and follower's states are determined using exteroceptive sensors such as cameras, GPS, IMU, lidar, ultra-wideband (UWB)~\cite{nguyen2019distance}, or the mixing of onboard sensors~\cite{scheidemann2024obstacle}. 

\begin{figure}
    \centering
    \includegraphics[width=0.99\linewidth]{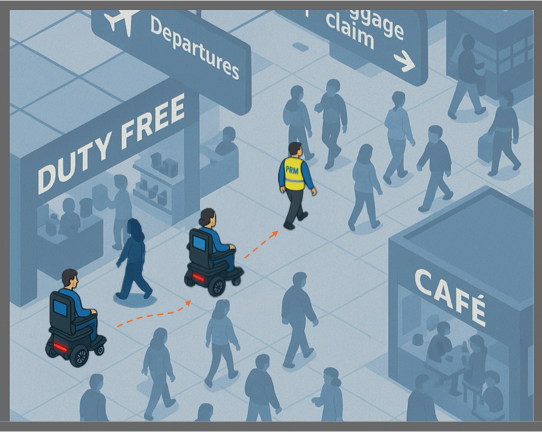}
    \caption{Illustration of the use case under study: one or multiple wheelchairs are to follow airport staff across dynamic and crowded environments, interacting with the proprietary obstacle avoidance, with controllers optimized for user experience.\\[-2em]}
    \label{fig:concept}
\end{figure}

Human-following, or follow-me tasks in general, are primarily classified into distance-position-based and perception-based approaches. In most position-based approaches, the follower tracks reference waypoints and trajectories by relying on the high accuracy relative positioning to the target~\cite{han2023leader}. Position-based systems can demand high computational resources when applied to nonlinear controllers or nonholonomic and aerial vehicles~\cite{nguyen2019distance}. Additionally, they may face challenges in achieving precise and continuous positioning in complex environments, depending on the capabilities of onboard sensors~\cite{lee2021uwb}. While in perception-based approaches, additional detection, tracking, and statistical learning methods calculate the geometrical states of the follower and the leader, and the control policy~\cite{tallamraju2019active}. 

In perception-based approaches, deep learning techniques have shown great potential as controllers for robotic systems. However, the implementation of these methods in real-world applications is limited due to the unstable approximation, safety concerns, and complexity~\cite{zare2024survey}. Recently, the use of end-to-end imitation learning (IL) has gained attention for developing controllers for autonomous driving~\cite{le2022survey}. In this process, desired actions are learned by 
training deep neural models on demonstration-based datasets to achieve performance comparable to human standards~\cite{osa2018algorithmic}.

In this work, we propose an end-to-end IL follow-me framework for assistive wheelchairs, designed to support individuals with limited mobility by autonomously following a human leader or another wheelchair. The vehicle tracks the target human using UWB devices, and its control actions are driven by an IL model trained on a demonstration-based dataset, allowing the wheelchair to mimic human movement patterns and decision-making. 

\begin{figure*}[t]
    \centering
    \includegraphics[width=0.95\linewidth]{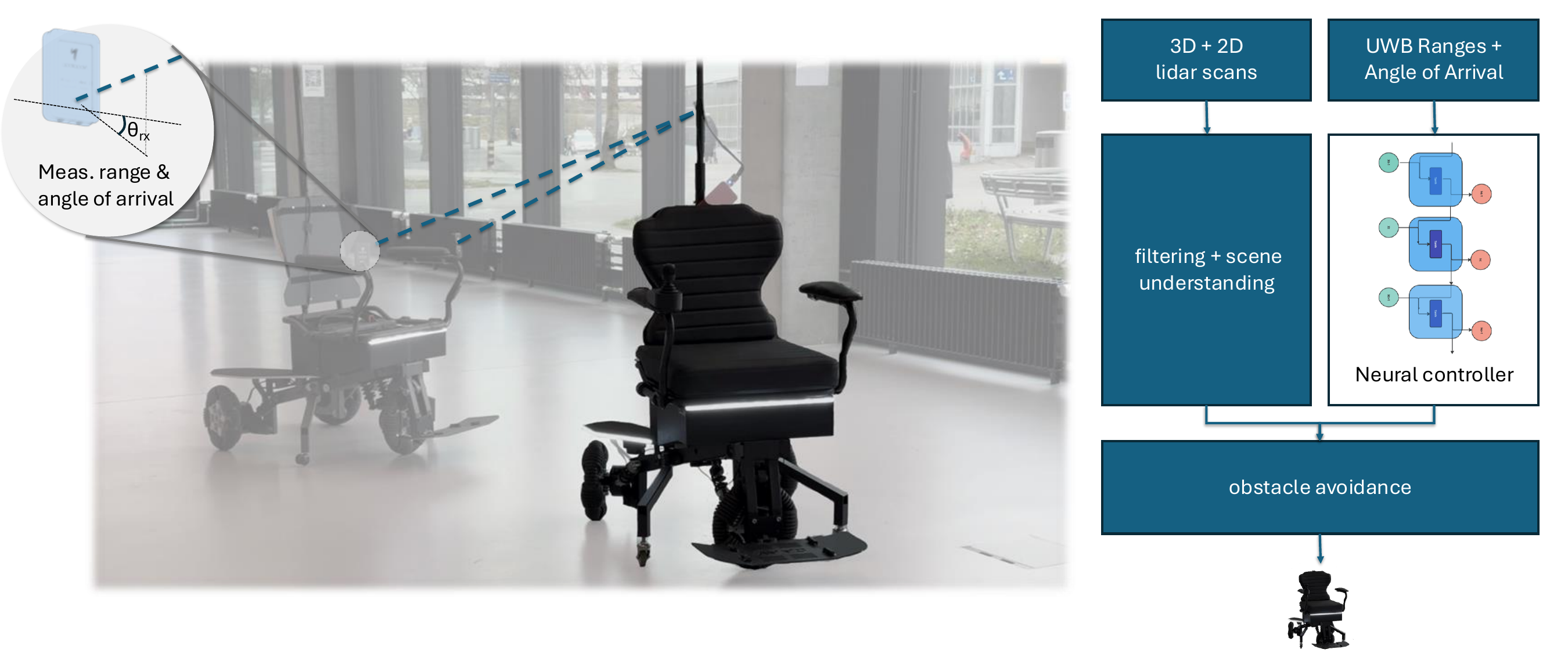}
    \caption{Left: DAAV Wheelchair equipped with UWB radio nodes for follow-me. Right, basic structure of the follow-me controller.}
    \label{fig:daav}
\end{figure*}

The demonstration-based dataset is based on UWB devices, which offer several key advantages, including low cost, high frequency, and centimeter-level accuracy, compared to complex and expensive vision-based and lidar sensors~\cite{salimpour2023exploiting}. Our learning framework incorporates various types of measurements derived from UWB, including ranging, angles, and their combination, enabling generalization and compatibility with different sensors. The objective of utilizing imitation learning in our proposed control model is to enhance safety, human-centered adaptability, and awareness. It also achieves natural motion and more intuitive and predictable navigation. 

The remainder of this paper is organized as follows. Section II provides an overview of existing approaches and applications in follow-me and leader-follower tasks, with a focus on UWB datasets and related work in imitation learning. Section III details the data acquisition process and model training setup. Experimental results are discussed in Section IV, with Section V concluding the paper.


\section{Background} \label{sec:related_work}

Existing Human-following applications typically rely on various types of sensors. Visual servoing has been widely applied to leader-follower scenarios~\cite{gupta2016novel,lin2020adaptive}. In~\cite{rollo2023followme}, the authors proposed a follow-me framework based on visual person detection and tracking using the ROS navigation stack, combined with additional control commands through hand gesture classification. Several studies have also investigated lidar sensors for precise localization and tracking of specific targets~\cite{eirale2025human}. A deep learning-based 3D object detection framework was employed in~\cite{park2024development} to detect and track the shape of human legs using 3D lidar data. 

Cameras and lidars are often costly, computationally demanding, and susceptible to losing track of the target in environments with similar targets. A potential solution to this challenge is emitter-receiver devices~\cite{jin2020robust}. For instance, in~\cite{pingali2018ultrasonic}, ultrasound receivers are used to determine the person’s precise position relative to the wheelchair, and~\cite{pingali2018ultrasonic} uses a Wi-Fi signal-based system to detect and track the person to follow.

UWB is also widely used in human-tracking and formation control applications~\cite{lee2021uwb,sarmento2022followme}. In~\cite{feng2018human}, as in many other common studies, three UWB anchors are mounted on the robot, and a virtual spring model is used to follow the tag’s position, which is calculated using the basic trilateration algorithm. In~\cite{xue2022uwb}, the authors proposed a side-by-side human-following method using UWB localization and adaptive prediction of the target person to adjust the robot’s motion parameters.

After detecting and localizing the target, autonomously navigating toward it can be challenging in real-world applications, especially when aiming to maintain smoother and safer trajectories.

Learning from expert demonstration often referred to as imitation learning, has recently emerged as a promising approach for end-to-end autonomous navigation~\cite{le2022survey}. For instance, a CNN-based deep imitation learning trajectory planner is proposed~\cite{chen2019deep}, using expert driving data from bird's-eye view map images. Also, in another study~\cite{codevilla2018end}, the authors introduced a conditional imitation learning, which maps expert commands together with camera image observations into steering angle and acceleration actions. However, limited research has explored the use of imitation learning for human-following tasks.

\begin{table}[b]
\centering
\caption{Comparison of different models' average performance over 10 iterations.}
\label{tab:model_comparison}
\begin{tabular}{@{}llccc@{}}
\toprule
\textbf{Model} & \textbf{Inputs} & \textbf{Outputs} & \textbf{Train MSE} & \textbf{Test MSE} \\ \hline
\multirow{3}{*}{MLP} & \( R_{1}, R_{2} \) & \( v_{x}, \omega_{z} \) & 0.0419 & 0.0429 \\ 
                    & \( A_{1}, A_{2} \) &  \( v_{x}, \omega_{z} \)  & 0.0551 & 0.0546 \\ 
                    & \( R_{1}, R_{2}, A_{1}, A_{2} \) &  \( v_{x}, \omega_{z} \)  & 0.0492 & 0.0509 \\ [1ex]
                    \hline
\multirow{3}{*}{LSTM} & \( R_{1}, R_{2} \) &  \( v_{x}, \omega_{z} \)  & 0.0267 & 0.0278 \\ 
                      & \( A_{1}, A_{2} \) &  \( v_{x}, \omega_{z} \)  & 0.0293 & 0.0306 \\ 
                      & \( R_{1}, R_{2}, A_{1}, A_{2} \) &  \( v_{x}, \omega_{z} \)  & \textbf{0.0154} & \textbf{0.0160} \\ [1ex] \hline
SVM                  & \( R_{1}, R_{2}, A_{1}, A_{2} \) &  \( v_{x}, \omega_{z} \)  & 0.0422 & 0.0437 \\ 
\bottomrule
\end{tabular}
\end{table}

\renewcommand{\arraystretch}{1.5}
\begin{table*}[t]
\caption{Comparison of different models' performance in different scenarios.}
\label{tab:model_comparison_experiments}
\centering
\resizebox{\textwidth}{!}{%
\begin{tabular}{@{}llccccccccc@{}}
\toprule
\multirow{2}{*}{\textbf{Model}} & \multirow{2}{*}{\textbf{Input data}} & \multicolumn{3}{c|}{\textbf{Experiment 1}} & \multicolumn{3}{c|}{\textbf{Experiment 2}} & \multicolumn{3}{c}{\textbf{Experiment 3}} \\
 &  & {MAE (degree)} & {Mean (m)} & {MAE (m)} & {MAE (degree)} & {Mean (m)} & {MAE (m)} & {MAE (degree)} & {Mean (m)} & {MAE (m)} \\
 \midrule
LSTM & \( R_{1}, R_{2} \) & 6.48 & 1.90 & 0.22 & 7.93 & 1.88 & 0.21 & - & - & - \\ 
                    LSTM & \( A_{1}, A_{2} \) & 6.81 & 3.12 & 1.1 & 8.30 & 2.26 & 0.22 & - & - & - \\ 
                    LSTM & \( R_{1}, R_{2}, A_{1}, A_{2} \) & \textbf{4.41} & 1.87 & 0.23 & \textbf{7.72} & 1.77 & 0.28 & \textbf{1.18} & 1.78 & 0.27 \\ 
\multirow{1}{*}{MLP} & \( R_{1}, R_{2}, A_{1}, A_{2} \) & 8.19 & 1.90 & \textbf{0.16} & 10.10 & 1.92 & \textbf{0.19} & 2.62 & 1.79 & \textbf{0.26} \\ 
\multirow{1}{*}{SVM}                  & \( R_{1}, R_{2}, A_{1}, A_{2} \) & 7.27 & 1.85 & 0.21 & 9.25 & 1.83 & 0.23 & 2.03 & 1.78 & 0.27 \\ 
\bottomrule
\end{tabular}
}

\end{table*}

\section{Methodology}
\subsection{DAAV's Holonomic Autonomous Wheelchair}

DAAV's autonomous wheelchair integrates safety mechanisms and mobility technologies to ensure secure and efficient transportation for passengers with reduced mobility in airport environments.

For situational awareness and obstacle avoidance, DAAV employs multiple 2D and 3D lidars, enabling the wheelchair to detect and navigate around obstacles in real-time, maintaining smooth and uninterrupted movement through crowded airport terminals. The lidar sensor data is fed to situational awareness and dynamic object tracking modules (Fig.~\ref{fig:daav}, right side).

The wheelchair's holonomic motion capability allows it to move freely in any direction, including lateral and rotational movements, through an omnidirectional wheelbase design. Complementing this hardware, proprietary obstacle avoidance algorithms process sensor data to dynamically adjust the wheelchair's path, ensuring safe navigation in complex and dynamic environments.

DAAV also incorporates UWB technology to support its "follow-me" mode, enabling the wheelchair to autonomously accompany a designated human leader. By leveraging radio-based ranging and angle measurements, UWB provides robust, centimeter-level localization in real-time, allowing the system to maintain responsive and intuitive tracking without reliance on visual inputs.

In this paper, DAAV serves as a real-world use case, having been deployed in operational environments with over 100 km of autonomous navigation logged across the Zurich and Vienna airports. Notably, its navigation and obstacle avoidance strategies are constrained by strict requirements for user comfort and overall user experience.

\subsection{Demonstration Phase}

The dataset was collected by a human teleoperating the wheelchair smoothly and safely following another human leader. During the demonstration phase, the leader followed a random trajectory, including moving forward, turning left and right, and stopping. The recorded dataset consists of 110 seconds of data from two UWB devices, including ranges $(r_1, r_2)$ and angles $(a_1, a_2)$ measurements to the target tag at a publishing frequency of 50 Hz. The distance to the leader ranged from 1 meter to 3.4 meters to cover a broader range of conditions. The action space is continuous, consisting of linear and angular velocity commands $(v, \omega)$, with the maximum linear velocity set to 1.2 m/s.

\subsection{Training Phase}

In this study, we evaluated three regression models for mapping continuous observations to velocity command actions. Given that our raw data contained sufficient information, we applied end-to-end learning without additional preprocessing. We compared Support Vector Machines (SVM), a single-layer Multi-Layer Perceptron (MLP), and a single-layer Long Short-Term Memory (LSTM) network as multi-output supervised learning regressors. Both the MLP and LSTM models, each with 32 units, were trained using the Mean Squared Error (MSE) loss function and optimized with the ADAM optimizer at a learning rate of 0.1. The LSTM model captures temporal dependencies by considering the historical performance of the leader over multiple timesteps. Given the high frequency of UWB data, we used a moving window of 2 seconds, corresponding to 100 sets of data.

\section{Experimental Results}
\subsection{Performance Evaluation}
The performance metrics of the models are presented in Table~\ref{tab:model_comparison}. In addition to utilizing both angle and range pairs as primary inputs in this study, we also evaluated the model performance using range-only and angle-only data. This analysis enables the generalization of the proposed framework to accommodate alternative sensor modalities. Based on the statistical results, the LSTM model demonstrated superior performance compared to the MLP and SVM models.

\begin{figure*}[hbtp]
    \centering
    \begin{minipage}[b]{0.32\textwidth}
        \centering
        \includegraphics[width=\textwidth]{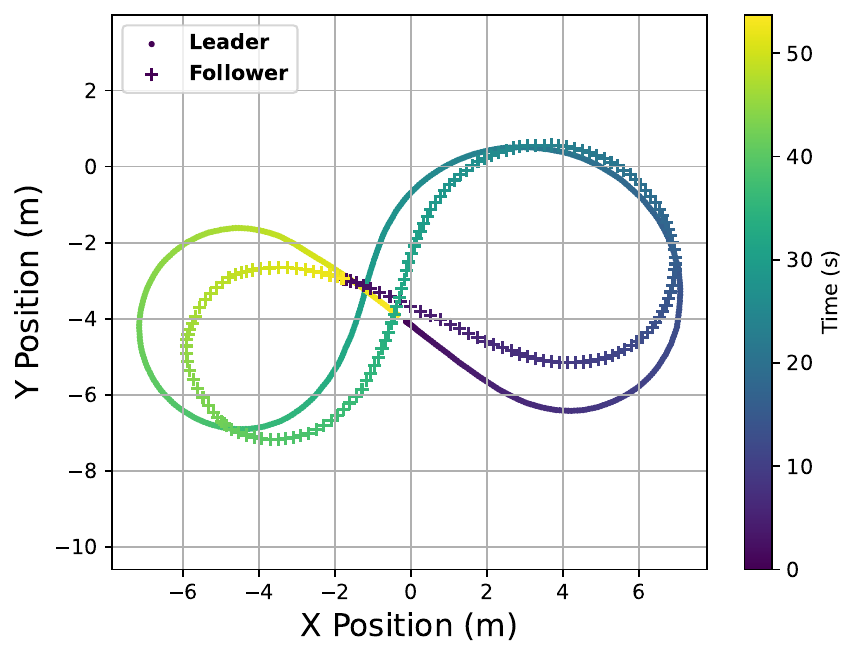}
    \end{minipage}
    \hfill
    \begin{minipage}[b]{0.32\textwidth}
        \centering
        \includegraphics[width=\textwidth]{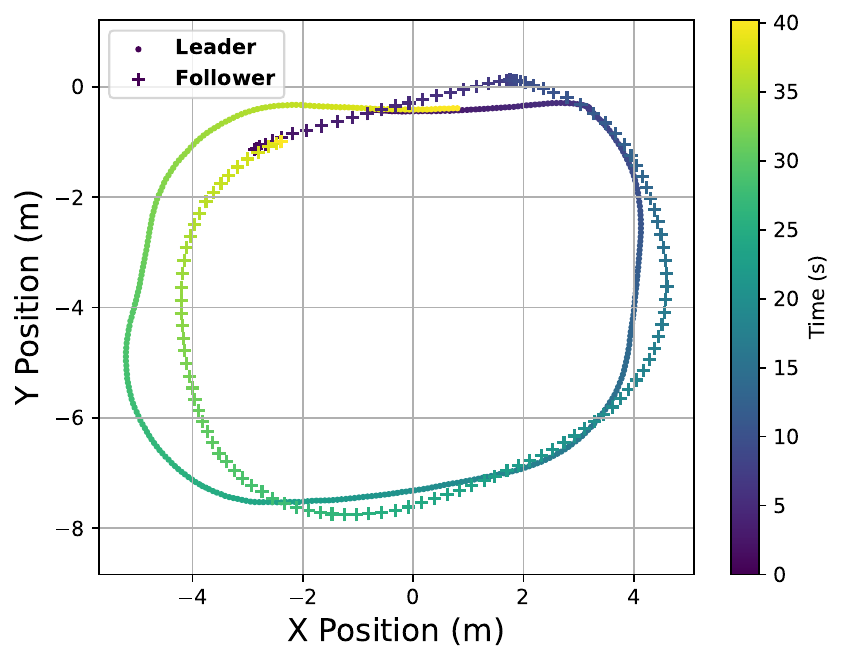}
    \end{minipage}
    \hfill
    \begin{minipage}[b]{0.32\textwidth}
        \centering
        \includegraphics[width=\textwidth]{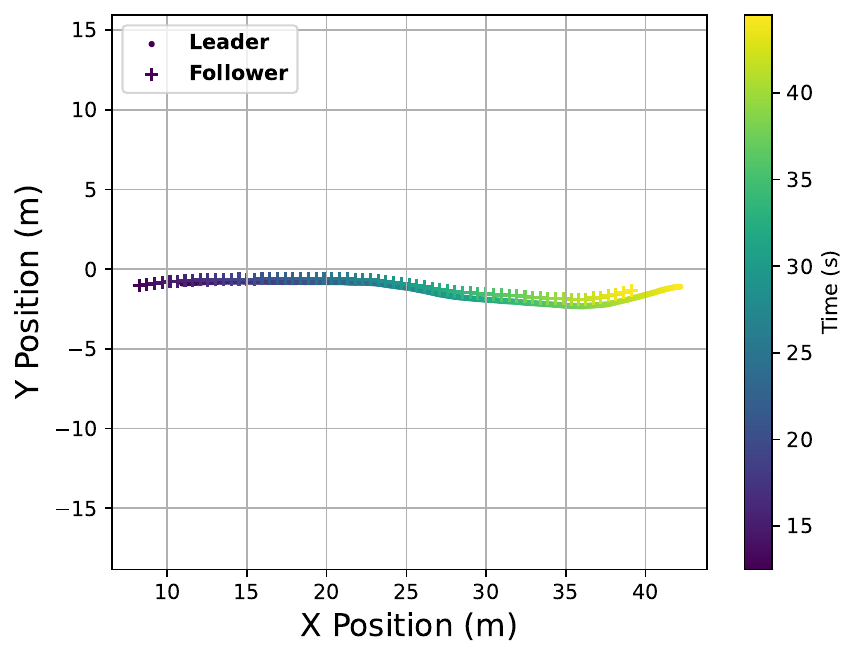}
    \end{minipage}

    \centering
    \begin{minipage}[b]{0.32\textwidth}
        \centering
        \includegraphics[width=\textwidth]{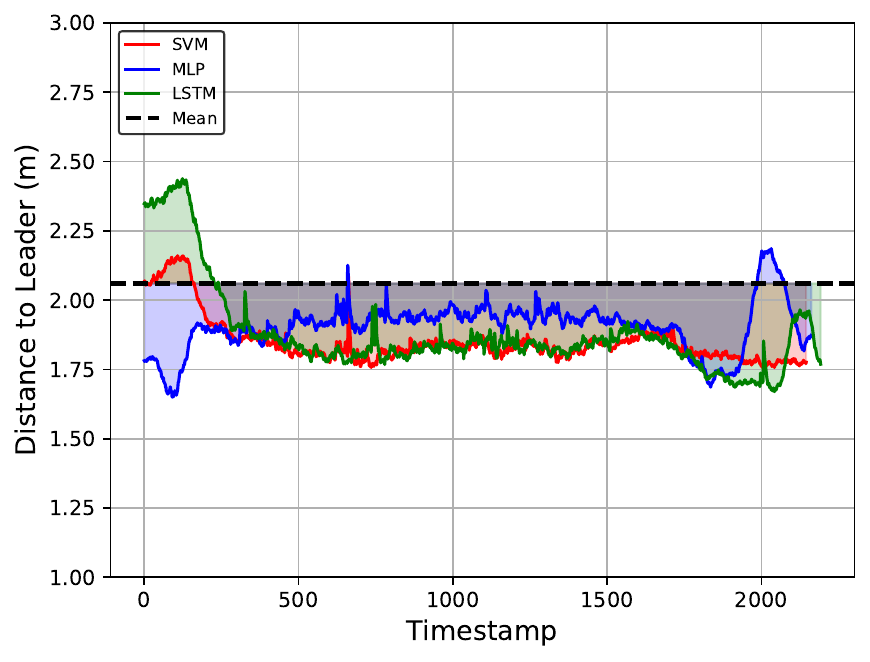}
        \subcaption{Experiment 1}
        \label{fig:dis_err_eight}
    \end{minipage}
    \hfill
    \begin{minipage}[b]{0.32\textwidth}
        \centering
        \includegraphics[width=\textwidth]{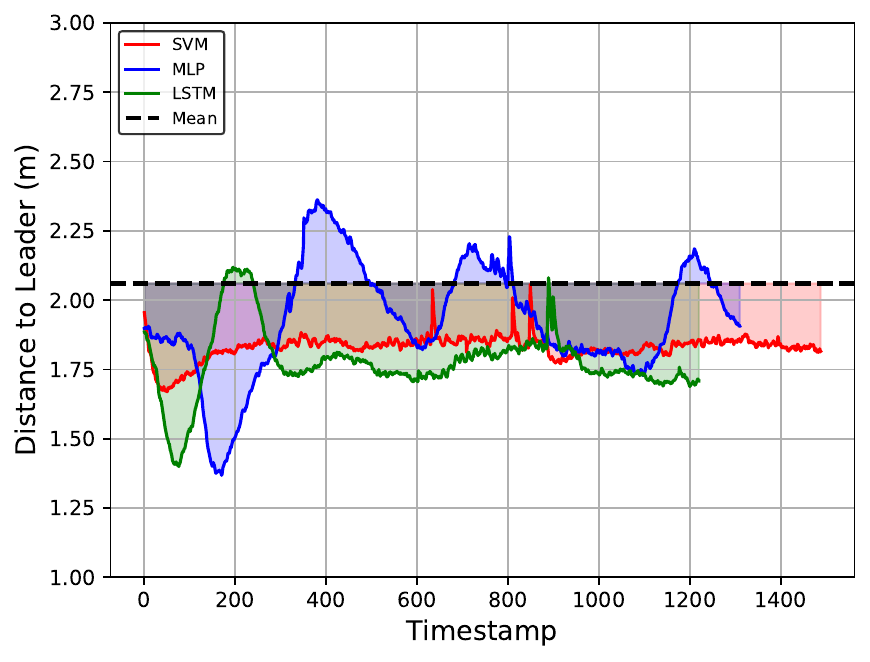}
        \subcaption{Experiment 2}
        \label{fig:dis_err_sq}
    \end{minipage}
    \hfill
    \begin{minipage}[b]{0.32\textwidth}
        \centering
        \includegraphics[width=\textwidth]{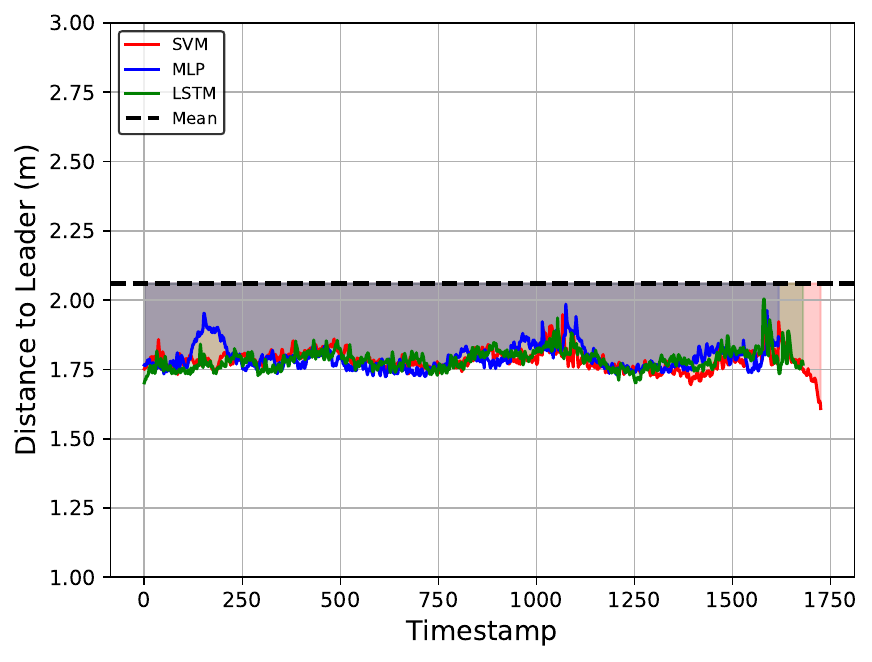}
        \subcaption{Experiment 3}
        \label{fig:dis_err_line}
    \end{minipage}
    \caption{Trajectory and distance between \textit{leader} and \textit{follower} across different experiments. The training data has a mean distance of $\sim2\,m$.}
    \label{fig:distance_error}
\end{figure*}

\begin{figure}[hbtp]
    \centering
    \begin{subfigure}{0.49\textwidth}
        \centering
        \includegraphics[width=\textwidth]{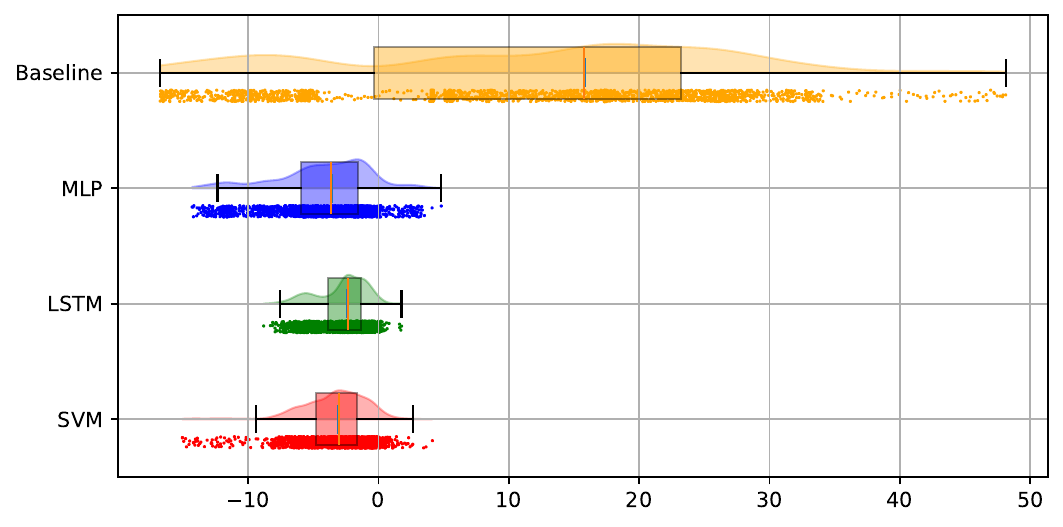}
        \caption{Angular offset ($deg$).}
        \label{fig:boxplot_angle}
    \end{subfigure}
    
    \begin{subfigure}{0.49\textwidth}
        \centering
        \includegraphics[width=\textwidth]{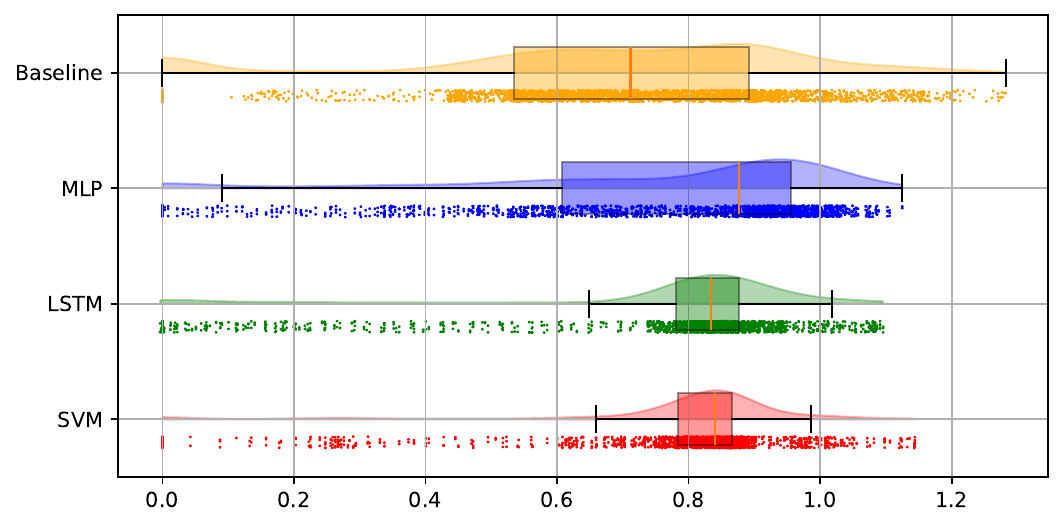}
        \caption{Linear speed ($m/s$).}
        \label{fig:boxplot_velocity}
    \end{subfigure}
    \caption{Comparison of the baseline model with three of the end-to-end deep learning models. The metrics represented here are proxies for the user experience and comfort. An angular offset near zero means the \textit{follower} wheelchair is pointing towards the \textit{leader}, meeting the user experience expectations. The linear speed mean value depends on the demonstrations, but the key here is that a low variance correlates to a \textit{smoother} ride, improving user comfort.}
    \label{fig:boxplots}
\end{figure}

\subsection{Real-World Validation}

We validated our framework in real-world follow-me navigation through three experiments. The performance of each model was evaluated based on the vehicle's ability to track the leader while maintaining a minimum angular deviation and a nominal distance consistent with the demonstration across different scenarios. Table~\ref{tab:model_comparison_experiments} summarizes the Mean Absolute Error (MAE) for angle offset and the mean distance between the follower chair and the target for models with various input configurations. The best angular tracking accuracy was obtained using the LSTM model with angle and range pairs as inputs, whereas the MLP model yielded the lowest mean absolute distance error. 

Figure~\ref{fig:distance_error} illustrates the trajectory of the follower chair tracking the leader across three distinct scenarios. Experiments 1 and 2 involved a broader range of movement patterns, while experiment 3 focused on maintaining a consistent following distance. The figure also presents the relative distance between the follower and leader over time, providing a detailed evaluation of tracking smoothness, a critical metric for real-world applications. Although the MLP model achieves the lowest average distance in table~\ref{tab:model_comparison_experiments}, it demonstrates notably poorer consistency relative to the LSTM and SVM models.

The angular difference and linear speed of the follower wheelchair relative to the leader serve as key indicators of ride smoothness and user comfort. Figure~\ref{fig:boxplots} presents a comparative analysis of these metrics across the three proposed models and a baseline model. These factors act as indicators for evaluating the overall user experience. As shown in the figure, all three models achieve smoother trajectories with significantly lower variance compared to the baseline, highlighting the benefit of imitation-based learning in enhancing ride quality.

Please see the attached multimedia material for videos of the experimental settings in crowded indoor environments.


\section{Conclusion}\label{sec:conclusion}

This paper presents an end-to-end imitation learning approach to follow-me functionality in micro-mobility applications. By leveraging real-world demonstrations and learning from expert trajectories, the proposed system learns to predict velocity commands directly from UWB-based spatial observations. We demonstrate significantly improved performance over the previous baseline, both in terms of tracking accuracy and stability, especially in dynamic and partially occluded environments.
The system has been successfully integrated into DAAV’s production codebase and validated in live robotic trials, highlighting its robustness and reliability in real-world conditions. 

\bibliographystyle{unsrt}
\bibliography{bibliography}

\end{document}